\documentclass[letterpaper, 10 pt, conference]{ieeeconf}  

\IEEEoverridecommandlockouts                              

\usepackage{graphics} 
    \DeclareGraphicsExtensions{.jpg,.pdf,.mps,.png} 
    \graphicspath{{fig/} {./}} 
\usepackage{epsfig} 
\usepackage{amsmath} 
\usepackage{amssymb}  
\usepackage{setspace} 
\usepackage{bm} 
\usepackage[flushmargin,hang]{footmisc} 
\usepackage{flushend} 

\newcommand{\fig}[1]{Fig.~\ref{#1}}

\newcommand{\eq}[1]{(\ref{#1})}

\def\epsgaiji#1{\leavevmode\kern-0.025zw\raise-.37zh\hbox{%
  \epsfile{file=#1,width=1.05zw}}\kern-0.025zw}
\newcommand{\MARU}[1]{{\ooalign{\hfil#1\/\hfil\crcr\raise.167ex\hbox{\mathhexbox20D}}}}

\newcommand{\sect}[1]{Sec.~\ref{#1}}

\usepackage{array}

\usepackage{siunitx} 
\usepackage{gensymb} 
\usepackage{multirow} 
\usepackage{caption}
\usepackage{subcaption}
\usepackage{colortbl} 
\usepackage{xcolor} 

\usepackage{pgfplots}
\pgfplotsset{compat=newest}
\usetikzlibrary{plotmarks}
\usetikzlibrary{arrows.meta}
\usepgfplotslibrary{patchplots}
\usepackage{xcolor}

\usepackage{grffile}
\pgfplotsset{plot coordinates/math parser=false}
\newlength\fwidth
\newlength\fheight

\usepackage{cite}

\title{\LARGE \bf
Structure from Motion-based Motion Estimation and\\
3D Reconstruction of Unknown Shaped Space Debris}

\author{Kentaro Uno$^{1*}$, Takehiro Matsuoka$^{1*}$, Akiyoshi Uchida$^{1}$ and Kazuya Yoshida$^{1}$
\thanks{$^{1}$K. Uno, T. Matsuoka, A. Uchida and K. Yoshida are with the Space Robotics Lab. (SRL) in Department of Aerospace Engineering, Graduate School of Engineering, Tohoku University, Sendai 980-8579, Japan.}%
\thanks{$^{*}$\textit{These authors contributed equally to this work. Corresponding author is Kentaro Uno. {\rm (E-mail: {\tt\small unoken@tohoku.ac.jp})}}}
}

\begin{document}

\maketitle
\thispagestyle{empty}
\pagestyle{empty}

\begin{abstract}
With the boost in the number of spacecraft launches in the current decades, the space debris problem is daily becoming significantly crucial. For sustainable space utilization, the continuous removal of space debris is the most severe problem for humanity. To maximize the reliability of the debris capture mission in orbit, accurate motion estimation of the target is essential. Space debris has lost its attitude and orbit control capabilities, and its shape is unknown due to the break.
This paper proposes the Structure from Motion-based algorithm to perform unknown shaped space debris motion estimation with limited resources, where only 2D images are required as input. The method then outputs the reconstructed shape of the unknown object and the relative pose trajectory between the target and the camera simultaneously, which are exploited to estimate the target's motion.
The method is quantitatively validated with the realistic image dataset generated by the microgravity experiment in a 2D air-floating testbed and 3D kinematic simulation.
\end{abstract}

\section{INTRODUCTION}
\label{sec:intro}


\subsection{Background}
\label{subsec:background}
\begin{figure}[t]
\centering
    \includegraphics[width=.95\linewidth]{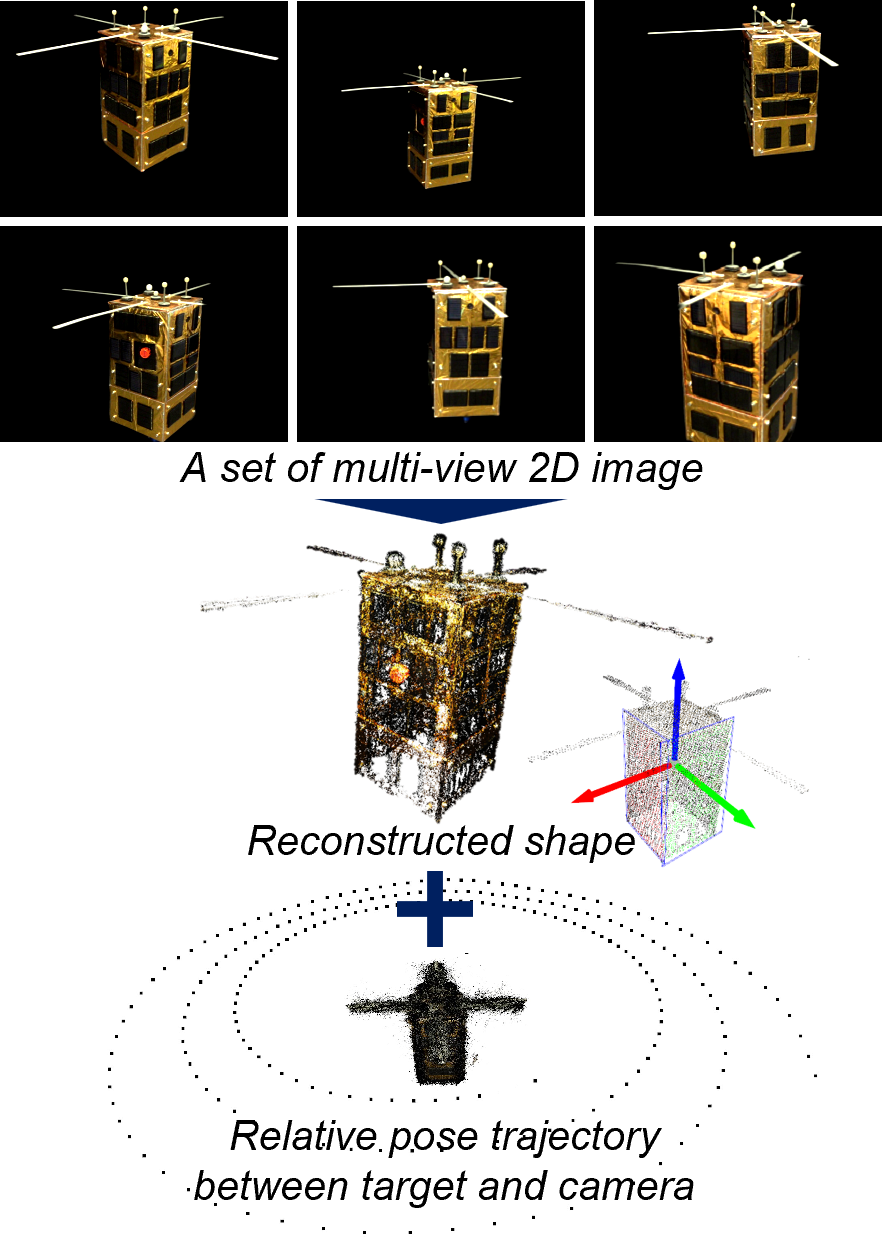}
    \caption{Concept of the Structure from Motion (SfM)-based motion estimation. SfM is usually applied to static objects with a moving camera; however, for the motion estimation of space objects, the stationary camera observes a free-flying target in orbit. Even so, SfM output is the same as the original use case, and the relative pose trajectory is exploited to estimate the target's motion parameters.}
    \label{fig:concept}
\end{figure}

Humanity has been conducting countless rocket launches to deploy many satellites and spacecraft in Low Earth Orbit (LEO) for various applications, e.g., wireless communication, observation, and habitabilization. In recent decades, the number of satellites in orbit has drastically increased due to the industrial domain's participation in space utilization. Notably, some companies and institutions have completed establishing such constellations into LEO (OneWeb: 648 satellites, Amazon Kuiper: 3,236 satellites, and SpaceX Starlink: 4,425 satellites)~\cite{osoro2023sustainability}. Cumulation in the number of space objects exponentially increases the risk of collision occurrence. Once such a conflict happens, broken spacecraft become out of control, making it impossible to reenter the atmosphere, becoming space debris. Such uncontrollability derives the higher risk of collision between the other functional satellites or debris, leading to the drastic self-propagation, called \textit{Kessler Syndrome}~\cite{kessler1978collision}. That smashes over many operational spacecraft and makes future orbital utilization almost irreversibly impossible.
To avoid such a catastrophe, while governmental institutions led the space mission to capture the in-orbit object, it should be mentioned that the dedicated startups' establishment and participation in such Active Debris Removal (ADR) (e.g., Astroscale in Japan, ClearSpace in Switzerland, and OrbitGuardians in the U.S.) 
prove more serious concerns also from the private domain.

Previous successful orbital missions include the rendezvous and docking by the chaser satellite with a robotic manipulator, which originated in the ETS-VII~\cite{ets7} program by the National Space Development Agency of Japan (NASDA, predecessor of JAXA) in 1998. The U.S. DARPA carried out a similar mission with the enhanced autonomy as ASTRO~\cite{darpaAstro} mission in 2007. In these experimental campaigns, the pre-designed target satellite, which is assumed to be space debris, is equipped with a dedicated grappling mechanism, and its attitude is fully controlled; however, indeed, many of the debris are \textit{uncooperative} --- meaning it has no such a docking connector but has uncontrolled tumbling behavior. Moreover, space debris may have lost its original shape due to collisions. Therefore, precisely predicting the target's motion and shape is essential before the capture sequence. Estimated motion parameters can then be fed back to the subsequent control for detumbling and capturing~\cite{uchida2024ISPARO} to maximize mission reliability.



\subsection{Related works}
\label{subsec:motivation}
To determine the space debris motion and shape, stereovision system is a straightforward approach; thus, many promising works are followed to date~\cite{sensorFusion,virtualStereoVision}; however, in the severely limited on-board computation and storage capacity, multi-frame 3D point cloud data handling and calculation tends to occupy the PC's memory.
It is also worth mentioning that the current Convolutional Neural Network (CNN)-based approach is showing encouraging results in spacecraft pose estimation~\cite{andrew,NNbasedPoseEst}. This type of method does not require inputting 3D data but needs a diverse 2D image dataset for training as a pre-process. For this, a virtually created dataset can be helpful for in-advance learning on the Earth to minimize the satellite servicer's internal computational load. Nonetheless, a technical gap still exists when applying it to unknown shaped objects.

\subsection{Contributions}
\label{subsec:contributions}
In this paper, we propose a method to estimate the motion of an orbiting object using \textit{Structure from Motion} (SfM), which does not depend on input 3D data and any pre-processes (a conceptual figure is shown as \fig{fig:concept}). Furthermore, the SfM-based method simultaneously performs a) the relative pose trajectory between the object and camera and b) 3D shape reconstruction of the target, inputting a set of multi-view 2D images of the target. This method is then helpful in predicting the motion of the unknown shaped space debris with the complex periodic motion. The major contributions of this paper are highlighted as follows: 
\begin{itemize}
    \item An autonomous sequence of pose estimation of the debris from SfM output is designed.
    \item Necessary equations to apply SfM to the floating object are presented.
    \item The method is validated with the image dataset having realistic microgravitational behaviors of the in-orbit object demonstrated in a 2D air floating testbed and 3D kinematic simulator.
\end{itemize}
To the best of the authors' knowledge, this paper is the first article to report an application of SfM to space debris motion estimation. The remaining part of this paper details the algorithmic process of the proposed method and presents experimental validations.

\section{METHOD}
\label{sec:method}

\subsection{Camera parameter calibration}
\begin{figure}[t]
  \centering
  \begin{minipage}[b]{0.45\linewidth}
    \centering
    \includegraphics[width=.9\linewidth]{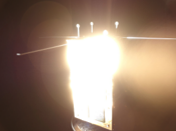}\\
    \footnotesize{(a) $f=1.8$.}
  \end{minipage}
  \begin{minipage}[b]{0.45\linewidth}
    \centering
    \includegraphics[width=.9\linewidth]{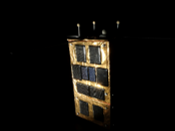}\\
    \footnotesize{(b) $f=11$.}
  \end{minipage}
  \caption{Pictures of the microsatellite mock-up exposed to the parallel intense light beam in a dark room with the different camera's aperture value: $f$. When $f$ is too small, (a) the reflective surface of the satellite becomes overexposed, resulting in unsuccessful SfM computation. However, (b) an appropriate f-number eliminates this issue, enabling proper feature detection.}
  \label{fig:overexposure}
\end{figure}
Spacecraft often use materials that readily reflect the illumination on their surfaces for thermal stability. Some specific relative orientation of the sun and the orbiting target causes the specular reflection, causing overexposure in some portions of the acquired images, which derives the crucial drop of the number of the natural features, and then miscalculation in SfM happens. \fig{fig:overexposure} depicts the results of an intense parallel light source applied to a satellite mock-up in a dark laboratory room. We confirmed that an appropriate setting of the camera's aperture value (f-number) effectively mitigates overexposure to achieve this objective. When using a low f-number, the camera's shutter speed should be set higher to mitigate motion blur. In the validation of this work, the f-number was simply set by the camera's auto-calibrating function or the default setting was used; however, manual calibration would be necessary in real-world applications.

\subsection{Background removal}
\begin{figure}[t]
\centering
    \includegraphics[width=.9\linewidth]{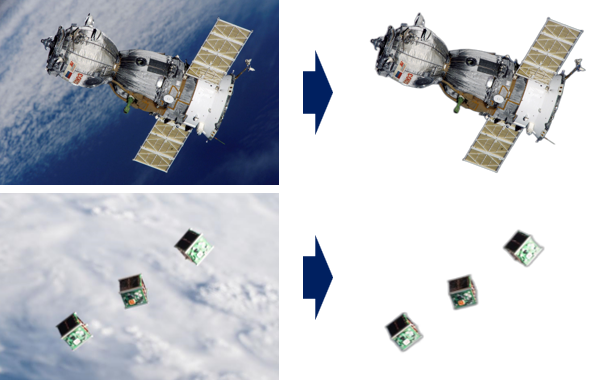}
    \caption{Possible in-orbit image examples (left) and the results of the background treatment where the target spacecraft was precisely extracted (right) (Images credit: \copyright NASA).}
    \label{fig:background_removal}
\end{figure}

In the case of space environments, depending on the camera's orientation, it is inevitable that Earth's surface or other planets appear in the background of the target image. Such background noise needs to be accurately removed to exclude inappropriate feature matching in the subsequent SfM process. Our algorithm employed rembg~\cite{rembg} for this process. Rembg performs foreground and background segmentation using the machine learning model $\mathrm{U^2}$-net~\cite{U2-Net}, and separates the background and the feature of interest using Python library for Alpha Matting~\cite{pymatting}, named $\text{pymatting}$. The result of background processing is exemplified in \fig{fig:background_removal}.

\subsection{Structure from Motion}
Structure from Motion (SfM)~\cite{SfM} is an algorithm that takes multiple images captured from different viewpoints as input and performs the following computations:
\begin{enumerate}
\item Detect and extract feature points from the input images.
\item Perform matching between the extracted feature points to calculate the relative positional and orientational movements between the object and the cameras.
\item Perform bundle adjustment to optimize the positions of the reconstructed shape points and the relative poses obtained in the previous step.
\end{enumerate}
In this algorithm, SfM processing is conducted using the open software. Based on the authors' trials of the representative open-source libraries with their default settings (e.g., feature descriptor matching schemes): a) OpenMVG~\cite{openmvg}, b) OpenSfM~\cite{opensfm}, and c) COLMAP~\cite{colmap}, a trade-off between the quality of the reconstruction (e.g., point cloud density) and computation time was confirmed. Library a) has the shortest runtime and the sparsest reconstruction, while library c) has the longest runtime and the densest reconstruction. Accordingly, this work employed b) OpenSfM, which offers intermediate quality and computational cost. The result obtained by applying SfM to the set of images after background processing is shown at the bottom of \fig{fig:concept}. In the center of the spiral trajectory, a point cloud representing the reconstructed 3D shape of the object is present, surrounded by the simultaneously estimated camera trajectories indicating their relative poses. The close-up view of the reconstruction is also shown in the same figure.

\subsection{Denoising and homogenization} \label{sect:denoise}
The three-dimensionally reconstructed point cloud obtained by SfM typically exhibits biased point cloud density. It may contain points unrelated to the object of interest, necessitating noise removal. In this algorithm, noise removal and point cloud homogenization are implemented using radius outlier removal and voxel down-sampling, provided by the Python library Open3D~\cite{open3d} for 3D point cloud processing. Radius outlier removal identifies points as outliers if the number of neighboring points within a specific radius sphere around each point is below a specified threshold. Voxelization-based down-sampling partitions the point cloud space into voxels of a specific size and creates a new point at the average position of points within each voxel, outputting the homogeneous point cloud. The result through these processes is shown in the middle of \fig{fig:concept} (denoised: colored point cloud, homogenized: smaller point cloud with frame next to the colored model).

\subsection{Target coordinate frame definition} \label{sect:targetFrame}
\begin{figure}[t]
  \centering
  \includegraphics[width=.95\linewidth]{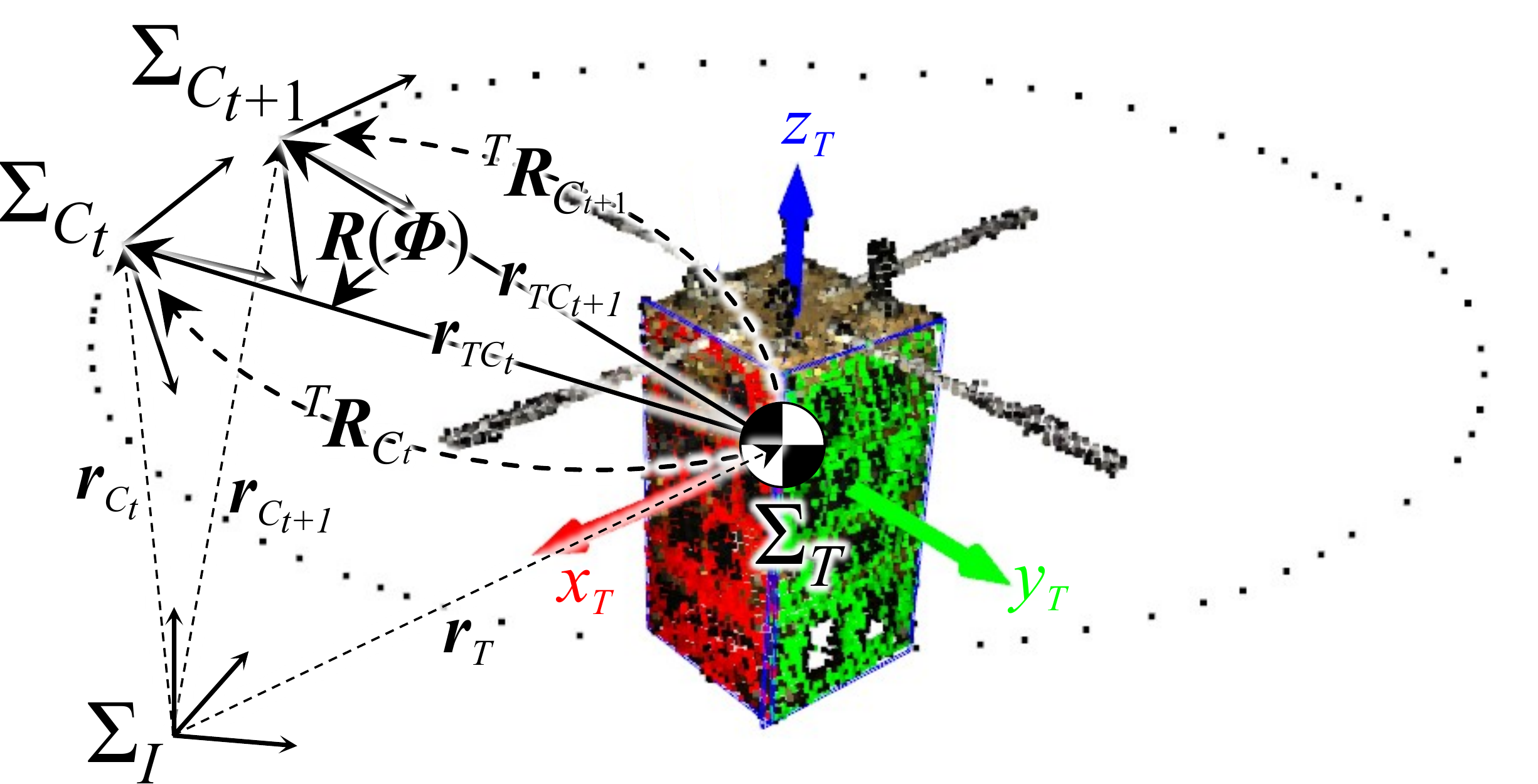}
  \caption{Principle for motion estimation from Structure from Motion output. Using the target frame $\Sigma_T$ defined by the user and the time sequential camera frame $\Sigma_{C_t}$ obtained by SfM, a sequence of the radius vectors (solid arrows) from the target to the moving camera are cumulatively computed. Target linear and angular velocity are estimated by measuring the change of the radius vector's norm and rotation, respectively.}\label{fig:motion_estimation_principal}
\end{figure}

In this study, the kinematic parameters are calculated with respect to a coordinate system $\Sigma_T$ fixed on the target body (see \fig{fig:motion_estimation_principal}). Although any option of the frame definition is available, adjusting the frame at the target's center of mass is the most practical approach. Considering that almost all space objects have relatively standard geometrical shapes (e.g., polygon: satellites, cylinder: fuel tank), and assuming the inertial property is homogeneous through the entire body, which means the center of geometry sufficiently matches the center of mass, the mass centroid is obtained by simple mathematics. 
In this paper, the micro CubeSat model is considered space debris in one representative case. Due to possible damage, it may not have a complete cubic shape, though some side planes are considered to be sufficiently flat. The technical tutorial designed in this work is detailed as follows. 
First, the object centroid is calculated. Given the denoised and homogenized point cloud, the average position of all points is regarded as the centroid, thus the origin of the fixed target coordinate system $\Sigma_T$. If the reconstructed model has the damaged part, broken sides are compensated by adding points with the same density to have a full cubic or cylindrical shape so that the geometrical center computation is more accurate. Second, plane detection is performed using pyRANSAC-3D~\cite{Mariga_pyRANSAC-3D_2022}, an open software based on RANSAC~\cite{ransac}. Through this process, we obtain the plane equations: $a_i x + b_i y + c_i z + d = 0$, where $i$ denotes the detected plane identifier. Finally, the corresponding normal vector of each plane $\bm{n}_i=[a_i, b_i, c_i]^\top$ can be used to define the representative axis of the polygon-shaped spacecraft, to which the axis of the target coordinate should be fixed, accordingly. In \fig{fig:motion_estimation_principal}, the red and green parts depict the detected planes from the 3D reconstruction.

\subsection{Motion parameter calculation}
\subsubsection{Translation}
Structure from Motion outputs a time sequence of the moving camera pose. Translational displacement is represented as the stretch of the norm of the moving radius from the target to the camera (see \fig{fig:motion_estimation_principal}). In the inertial coordinate system $\Sigma_I$, calculating the moving radius vector $\bm{r}${\footnotesize$_{\!TC_t}$} $ = $ $\bm{r}${\footnotesize$_{\!C_t}$} $-$ $\bm{r}${\footnotesize$_{\!T}$} at each time period $t$, change of the norm of the moving radius $L$ is then given as follows.
\begin{eqnarray}
L = \| \bm{r}{\scriptstyle _{\!TC_{t+1}}} \| - \| \bm{r}{\scriptstyle _{\!TC_{t}}} \| 
  = \| \bm{r}{\scriptstyle _{\!C_{t+1}}} - \bm{r}{\scriptstyle _{\!T}} \| -  \| \bm{r}{\scriptstyle _{\!C_{t}}} - \bm{r}{\scriptstyle _{\!T}} \|
\end{eqnarray}
Then, the translational velocity $\bm{v}$ is estimated with the division of $L$ by time step $t$, which is equivalent to the reciprocal of the camera frame rate.
\begin{eqnarray}
    \bm{v} = c~\frac{L}{t}~\bm{e}_d 
    \label{eq:velocity_estimation}
\end{eqnarray}
in which, $c$ is scale coefficient. It is important to note that this is necessary because there is no reference scale in the SfM space. Thus, a representative scale factor must be predefined for this method to adjust the estimated parameter in the real world's dimensions. For the application in this study, the satellite's edge or solar panel length and diameter of the antenna can be used. To describe the translation as a vector value, unit direction vector $\bm{e}_d$ is multiplied, defined in the desired reference frame, such as $\Sigma_T$.

\subsubsection{Rotation}
In principle, rotation is represented as the angular motion of the moving radius $\bm{r}${\footnotesize$_{\!TC_t}$} in SfM output (\fig{fig:motion_estimation_principal}). We need to pay attention to the fact that the angular movement of the camera pose in time is oppositely shown to the desired satellite's rotation in the space of SfM output. Thus, letting the rotation matrix $\bm{R}(\bm{\Phi})$ that enables the rotation $\bm{\Phi}$, which is the parameter estimated, its inverse operator maps the rotational transform of the camera frame in target fixed frame at $t$ to the next one at $t+1$. This is described in the following equation, where {\footnotesize$^{A\!}$}$\bm{R}${\footnotesize$_{B}$} $\in \mathbb{R}^{3\times3}$ is the rotation matrix from coordinate $\Sigma_A$ to coordinate $\Sigma_B$.
\begin{eqnarray}
    ^{\scriptscriptstyle {T\!}}\bm{R}_{\scriptscriptstyle C_{t+1}} = \bm{R}^{\scriptscriptstyle -1} \cdot ^{\scriptscriptstyle  {T\!\!\!}}\bm{R}_{\scriptscriptstyle C_{t}}
\end{eqnarray}
From this, the following equation can express the attitude change of the satellite from time $t$ to $t+1$ in $\Sigma_T$.
\begin{eqnarray}
    \bm{R}(\bm{\Phi}) 
    = \, ^{\scriptscriptstyle {T\!}}\bm{R}_{\scriptscriptstyle C_{t}} \cdot ^{\scriptscriptstyle {T\!\!\!}}\bm{R}_{\scriptscriptstyle C_{t+1}}^{\scriptscriptstyle {-1}}
    = \, ^{\scriptscriptstyle {T\!}}\bm{R}_{\scriptscriptstyle I} \cdot ^{\scriptscriptstyle {I\!\!\!}}\bm{R}_{\scriptscriptstyle C_{t}} \cdot \left[ ^{\scriptscriptstyle {T\!}}\bm{R}_{\scriptscriptstyle I} \cdot ^{\scriptscriptstyle {I\!\!\!}}\bm{R}_{\scriptscriptstyle C_{t+1}} \right]^{\scriptscriptstyle {-1}}
\end{eqnarray}
In this equation, all the rotation matrices on the left side are obtainable in SfM output, then the desired rotation angles $\bm{\Phi}$ is extracted out from $\bm{R}(\bm{\Phi})$.

\section{VALIDATIONS}
We conducted an experiment and a simulation to obtain the two- and three-dimensional expected motion in microgravity.

\begin{figure}[t]
  \centering
  \includegraphics[width=.8\linewidth]{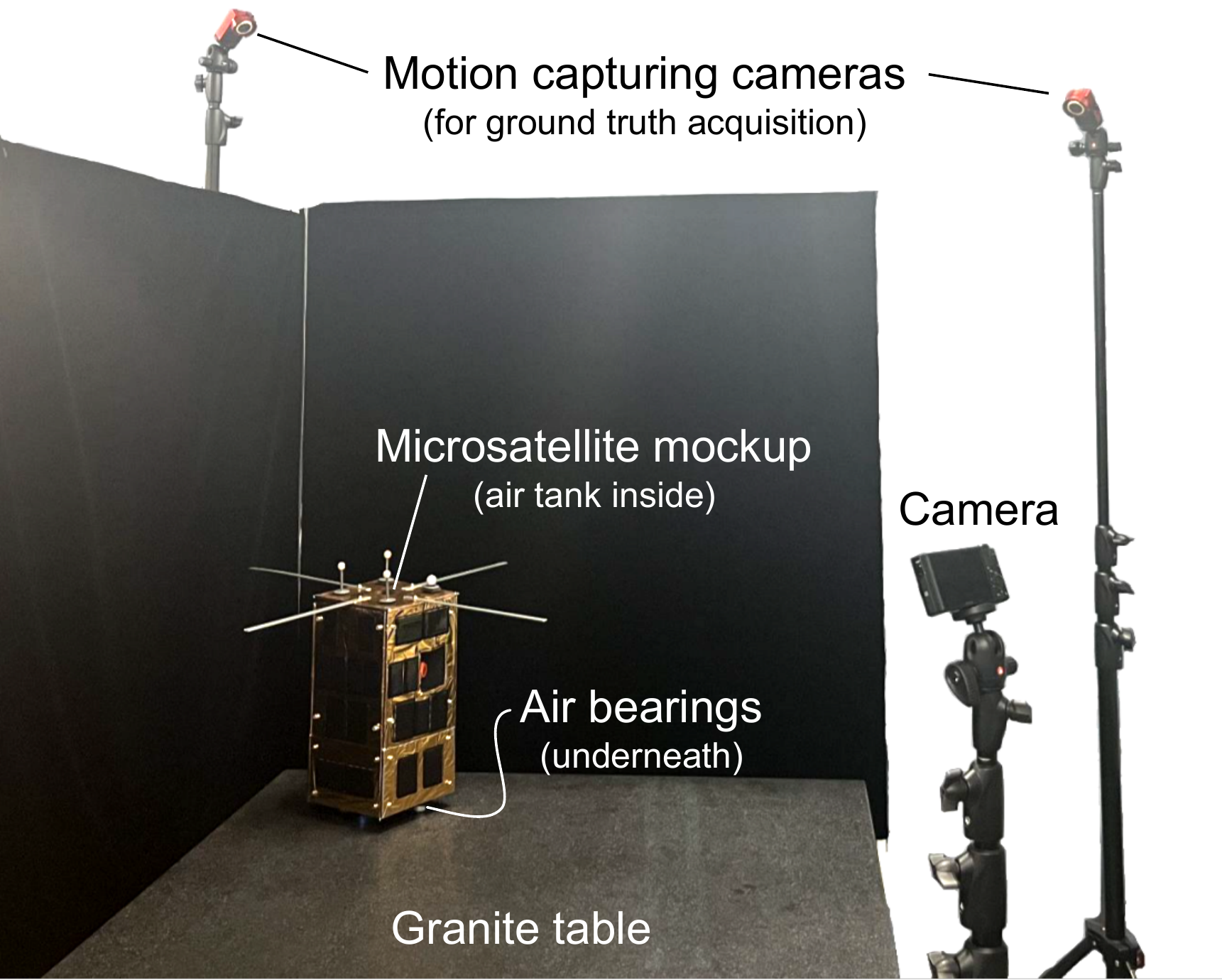}
  \caption{Experimental setup with air-floating testbed to emulate two-dimensional microgravity motions. Motion capture system was used to get the ground truth for comparison.}\label{fig:air_floating_testbed}
\end{figure}
\begin{figure}[t]
  \centering
  \begin{minipage}[b]{0.45\linewidth}
    \centering
    \includegraphics[keepaspectratio,height=.9\linewidth]{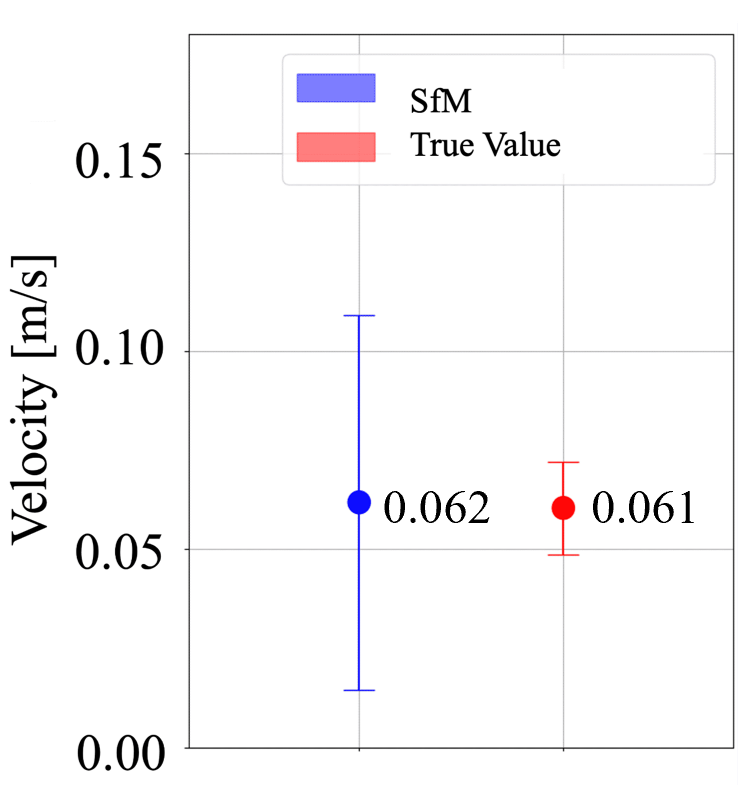}\\
    \footnotesize{(a) Linear velocity.}
  \end{minipage}
  \begin{minipage}[b]{0.45\linewidth}
    \centering
    \includegraphics[keepaspectratio,height=.9\linewidth]{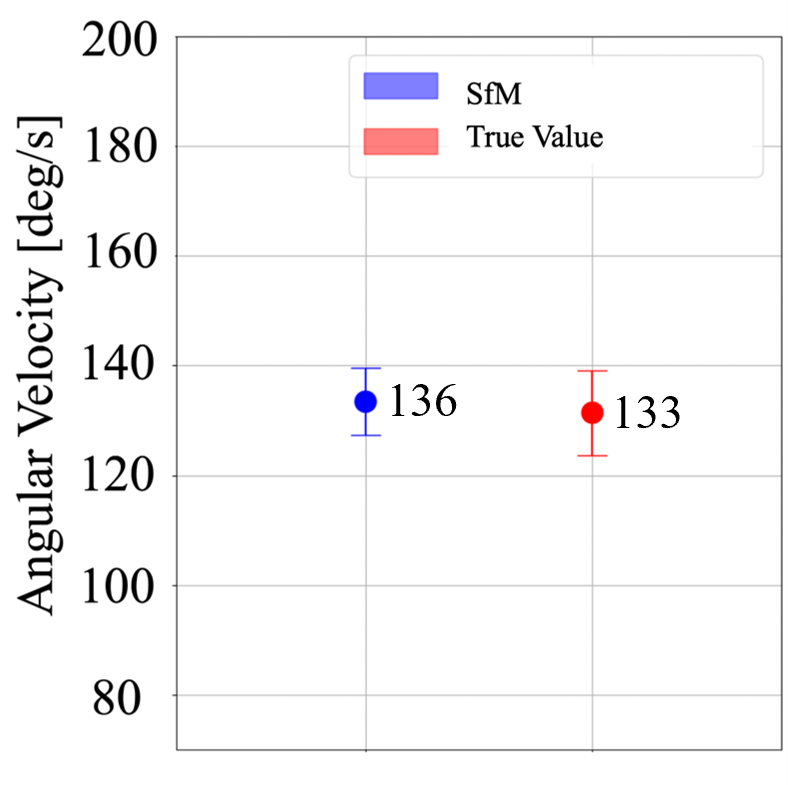}\\
    \footnotesize{(b) Angular speed.}
  \end{minipage}
  \caption{Comparison between the estimation and true value in the kinematic parameters of the two-dimensional uniform motion expected in orbit. 
  Dots are the mean values for \SI{7}{sec} of motion, and the error bars are their standard deviations.}
  \label{fig:comparison of vel and acc for rot and trans}
\end{figure}
\begin{figure*}[t]
  \centering
  \includegraphics[width=\linewidth]{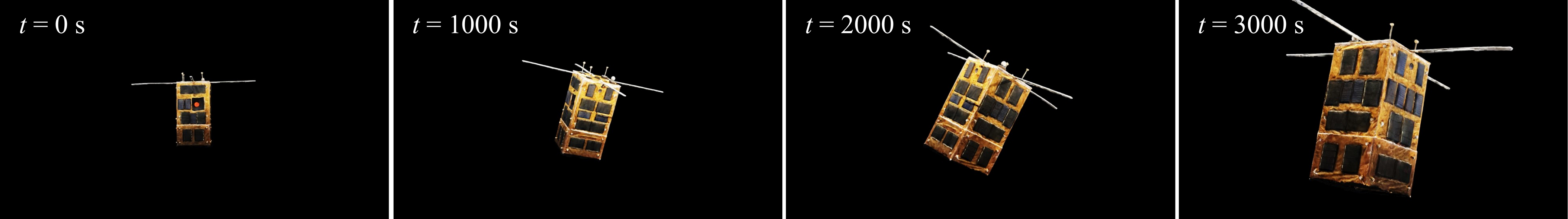}
  \caption{Sinusoidal motion of the in-orbit satellite generated by the simulation. For a servicer satellite approaching the target, an area of the target appearance (field of interest) grows over time. Resolution of the images: $1920 \times 1080$~pixels.}\label{fig:nerf_data_set}
\end{figure*}
\begin{figure}[t]
  \centering
  \includegraphics[width=\linewidth]{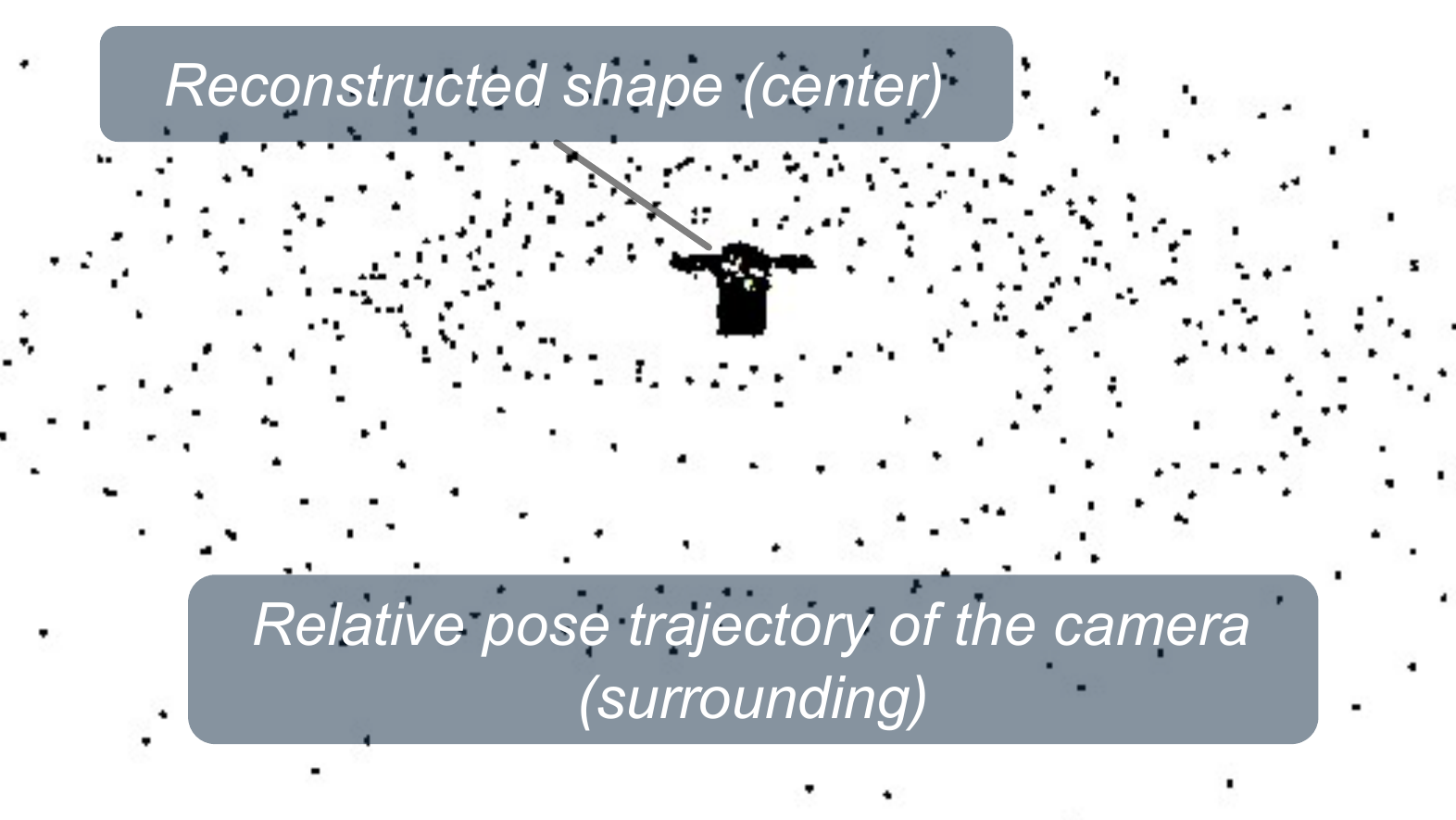}
  \caption{Capture of the result of SfM applied to the target satellite model in a 3D sinusoidal tumbling.}\label{fig:nerf_data_set_sfm_result}
\end{figure}

\subsection{Two-dimensional motion estimation}\label{sec:results_of_2d}
Among variable solutions to emulate microgravitational behavior~\cite{uno2024lower}, in this work, an air-floating testbed is employed. With this apparatus, pressurized air is injected downward through the air bearings attached underneath the satellite mock-up, realizing the frictionless motion on a flat surface (see \fig{fig:air_floating_testbed}). This type of testing is useful to simulate the single-axis rotation and planar translation in microgravity precisely. Such a motion is observed in actual satellites to stabilize their altitude.

In this setup, the mock-up is given an initial force that causes the $z$-axis rotation and $xy$-planar translation. The experiment was performed under normal illumination conditions in a laboratory room, and the f-number was set by the camera's autoexposure function. The motions were recorded by a single RGB camera: Sony DSC-RX100M5A with a frame rate of \SI{120}{fps} and a resolution of $1920 \times 1080$~pixels. Each first frame was extracted every four frames from the recorded video; thus, a set of \SI{30}{fps} images are processed to the algorithm. The specifications of the used PC are CPU: Intel Core i7-11700K, RAM: \SI{64}{GB} memory, and GPU: NVIDIA GeForce RTX 3090/24GB. Motion capture system was used to gain the ground truth for evaluation.

Throughout \SI{7}{sec} in total time of the motion, its linear and angular components were uniform thanks to the frictionless experiment. SfM output of such a 2D constant velocity motion is showcased at the bottom of \fig{fig:concept}. Estimation results are summarized in \fig{fig:comparison of vel and acc for rot and trans}. As a scale factor $c$ in Eq.~\eq{eq:velocity_estimation}, \SI{60}{mm} of the solar panel's longer edge was used for estimation. The graphs compare the mean value of linear and angular speeds between estimation and ground truth, showing 5\% and 2.3\% relative errors, respectively. These error deviations should include mitigating the speed due to non-zero friction on the table and atmospheric viscosity in the room. Notably, the proposed method is applicable even when the target is not rotating but only translating. In this case, only the partial structure of the target facing the camera is reconstructed unless the camera moves around the object. Nonetheless, since the typical motion of space debris includes periodic angular motion, stationary observation by the camera is sufficient.

\begin{figure*}[t]
  \centering
  \begin{minipage}[b]{0.325\linewidth}
    \centering
    \includegraphics[keepaspectratio,width=1.05\linewidth]{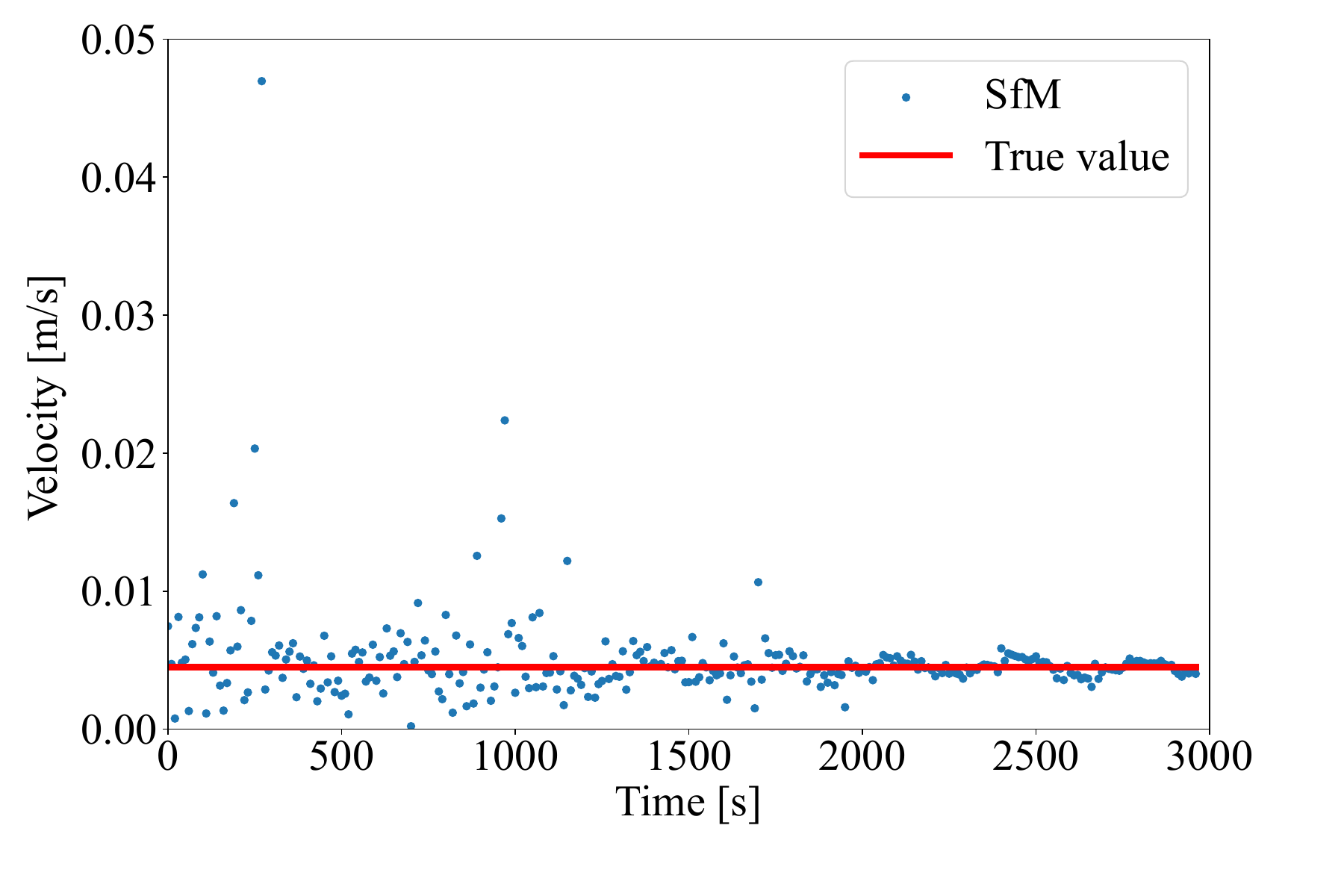}\\
    \footnotesize{(a) Linear velocity \\(RMSE: \SI{0.0002}{m/s}).}
  \end{minipage}
  \begin{minipage}[b]{0.325\linewidth}
    \centering
    \includegraphics[keepaspectratio,width=1.05\linewidth]{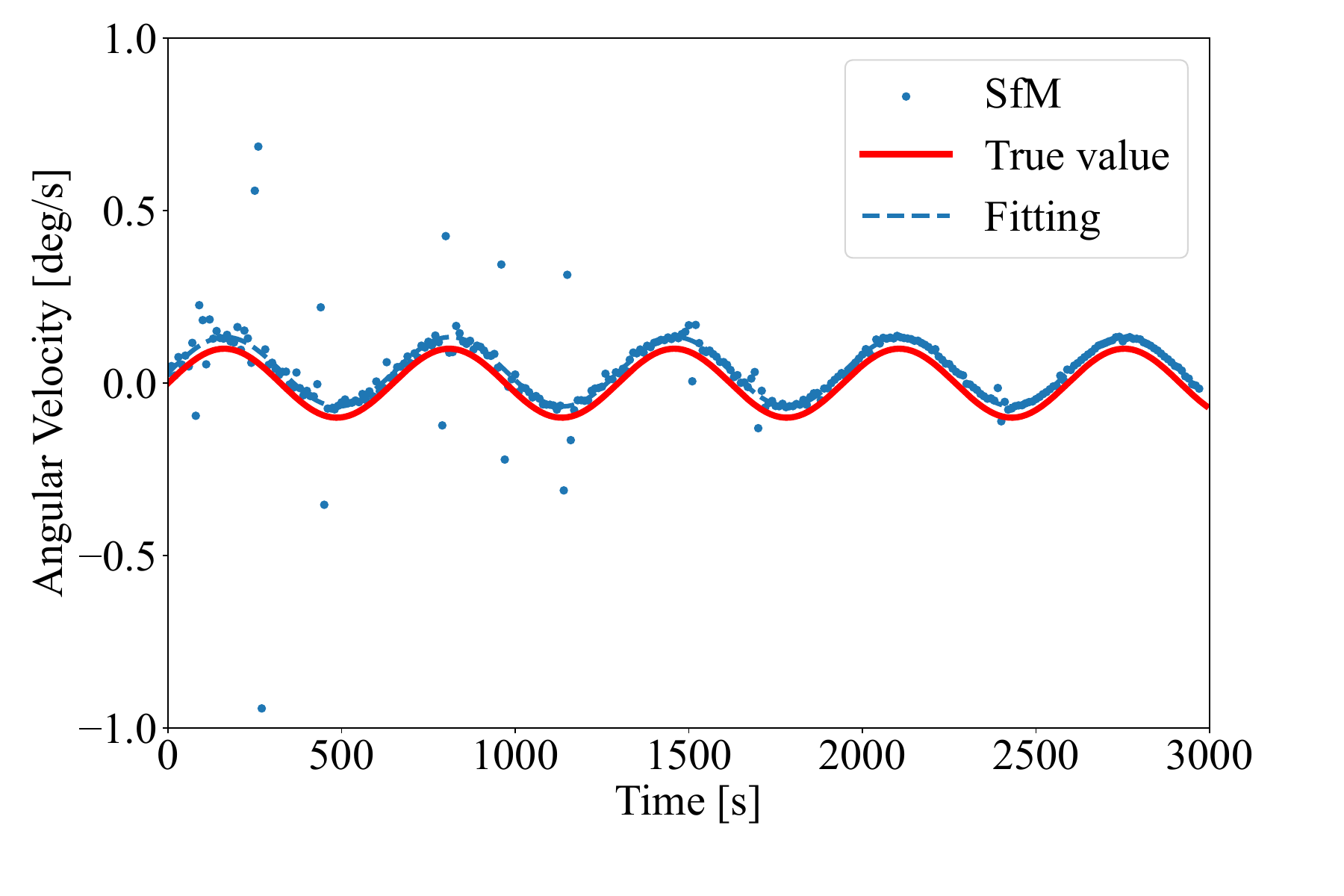}\\
    \footnotesize{(b) $x$-axis angular velocity (RMSE: \SI{0.0056}{deg/s}, relative error of period: 0.043\%)}
  \end{minipage}
  \begin{minipage}[b]{0.325\linewidth}
    \centering
    \includegraphics[keepaspectratio,width=1.05\linewidth]{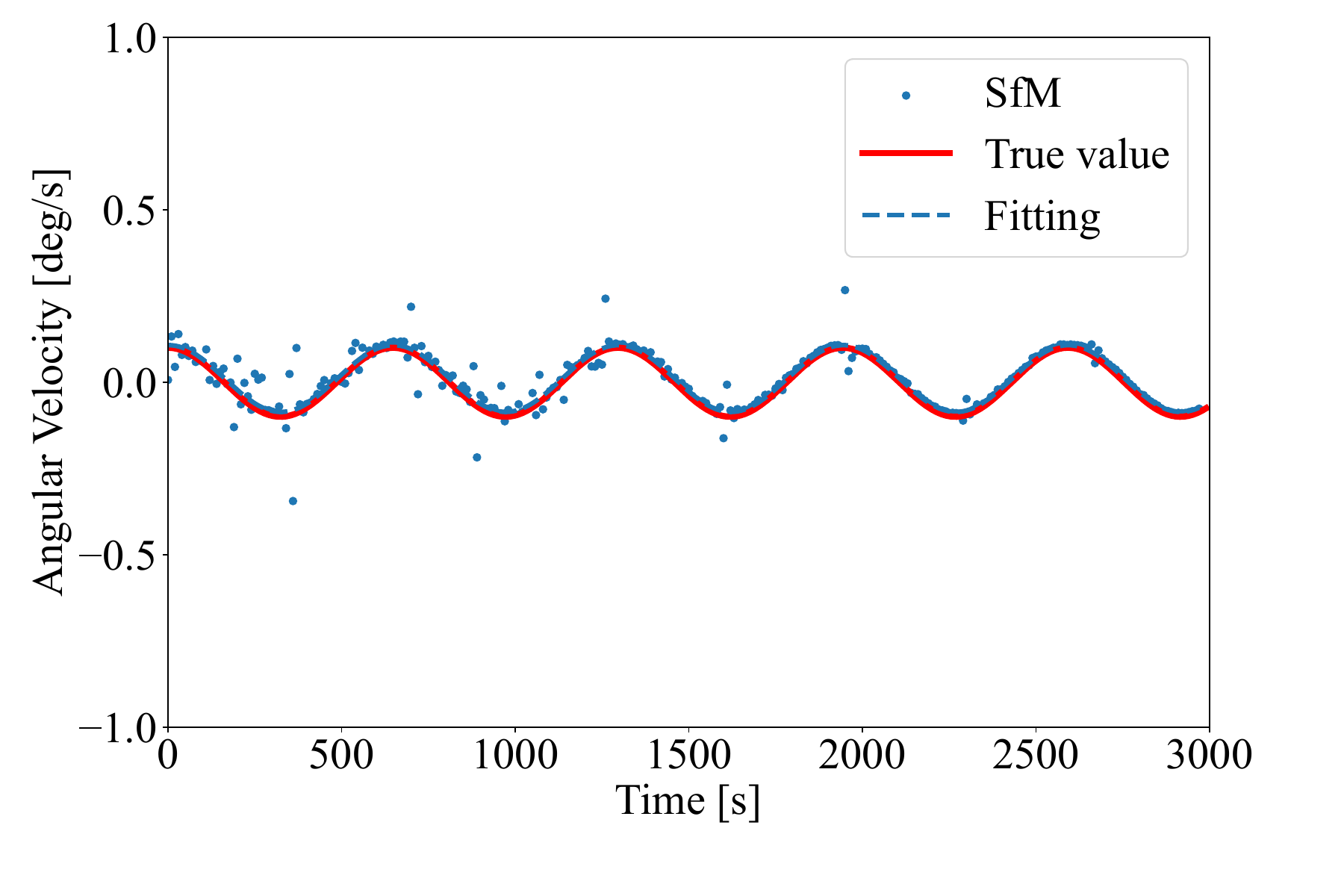}\\
    \footnotesize{(c) $y$-axis angular velocity (RMSE: \SI{0.0021}{deg/s}, relative error of period: 0.006\%)}
  \end{minipage}
  \caption{Time history of the estimated kinematic parameters (blue dots) compared with the true values (red lines). Periodic behavior in the rotation is reproduced by sine curve fitting from discrete estimated points (blue dashed lines).}
  \label{fig:nerf_data_set_motion_estimation_result_analysis}
\end{figure*}
\begin{figure}[t]
  \centering
  \includegraphics[width=.7\linewidth]{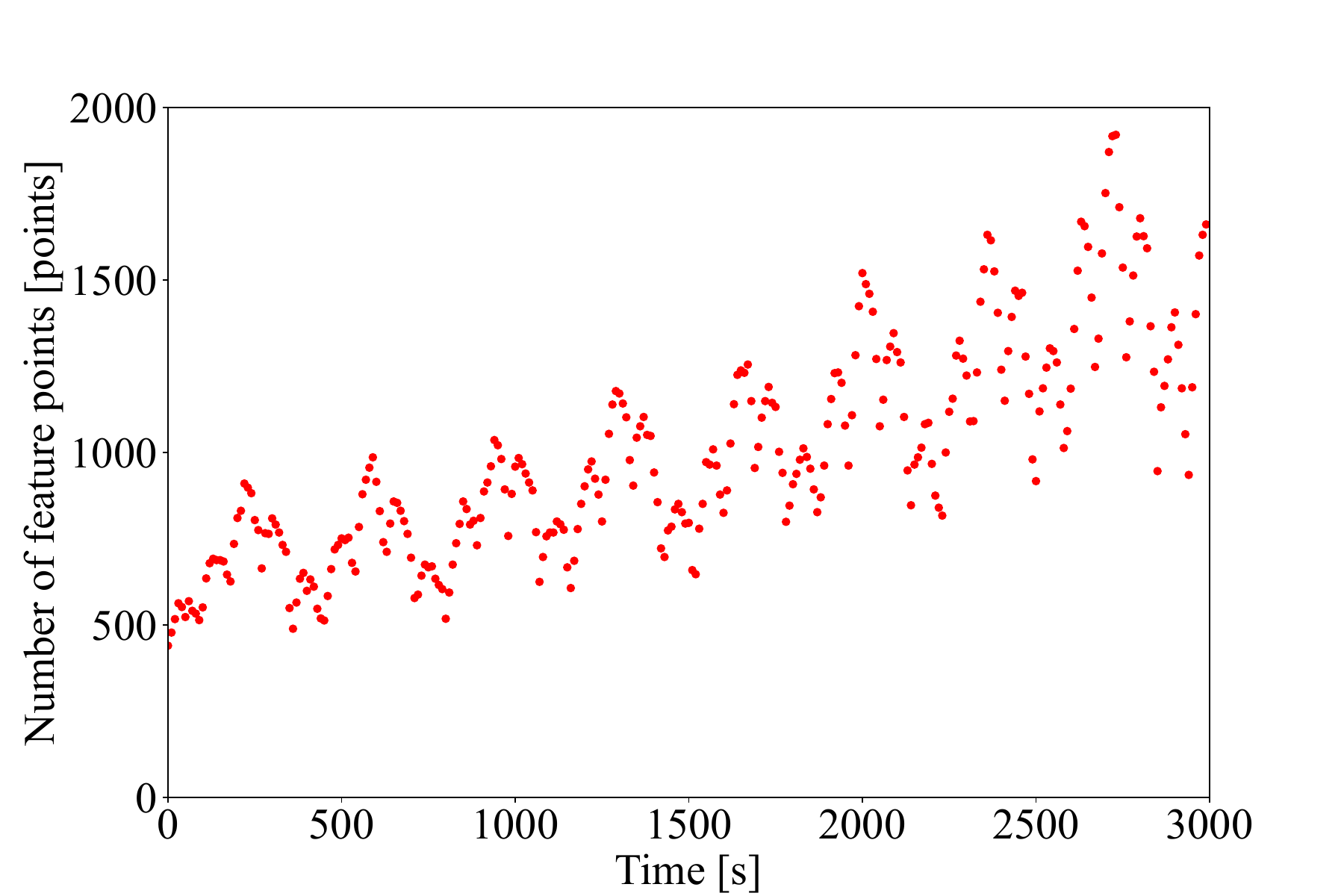}
  \caption{Change of the detected feature points used SfM calculation in time.}\label{fig:feature_point_increasing}
\end{figure}

\subsection{Three-dimensional motion estimation}\label{sec:results_of_3d}
Space debris can be in three-dimensional sinusoidal motion in orbit. While it is complicated to demonstrate such a motion under the Earth's gravity environment, simulation enables us to make the conditions closer to space. Therefore, we reproduced the microgravity motion in a virtual space by kinematic simulation. A 3D digital twin of the satellite mock-up was created using the 3D scenery representation tool NeRF~\cite{nerf}. This digital model is used in the 3D computer graphic software Blender to reproduce the spatial motion and obtain images.
The motion was calculated with reference to the satellite fixed coordinate system. The kinematic simulation was performed by solving Euler's motion of equation: $ \bm{N} = \bm{I} \Dot{\bm{\omega}} + \bm{\omega} \times \bm{I \omega}$ with the fourth-order Runge-Kutta integration, giving an inertial property and the initial condition. In this simulation, principal (diagonal) components of the inertia matrix $I_{xx}, I_{yy}, I_{zz}$ were set as $I_{xx}=I_{yy}=0.47$ and $I_{zz}=0.02$, and the initial linear and angular velocity is given as $\bm{v_0}=[0.0045,~0,~0]^\top$~[m/s] and $\bm{\omega}_0=[0,~0.1,~1.0]^\top$~[deg/s], where $x$-axis is toward the camera and $z$-axis is along to the bottom-top direction of the satellite at $t=0$. This motion is observed by the virtual camera in Blender. A default camera parameter in Blender was selected. Snapshots of the simulated sinusoidal behavior are exhibited in \fig{fig:nerf_data_set}.

SfM was then applied to this \SI{50}{min} of motion by inputting a set of images taken at every \SI{10}{sec}, meaning 300 frames in total. The image resolution and the used PC are the same as in the last section. The output of SfM is shown in \fig{fig:nerf_data_set_sfm_result}, where the moving radius from the reconstructed target to each camera trajectory point represents the translation by its norm and rotation by its change of direction. Estimated linear and angular velocity around the $x_T$- and $y_T$-axes relative to the fixed frame $\Sigma_T$ are compared with the true value in time (see \fig{fig:nerf_data_set_motion_estimation_result_analysis}). Root mean square error (RMSE) showed satisfactory accuracy of the estimation: less than \SI{1}{mm/s} or \SI{0.005}{deg/s}. Also, the period error in the rotational components is negligible. 

\subsection{Discussions}
In both 2D and 3D motion estimation, linear motion appears to have more noise compared to rotational motion. This is likely because linear motion errors are more sensitively affected by errors occurring in the centroid estimation, as explained in \sect{sect:denoise}--E. In 3D motion estimation, we observe few significant errors in estimation; in the macro viewpoint, the number of such errors is higher in the beginning phase (see \fig{fig:nerf_data_set_motion_estimation_result_analysis}(b)); in the micro viewpoint, outliers can be found mainly at the peak of the periodic motion (see \fig{fig:nerf_data_set_motion_estimation_result_analysis}(c)).
Such an error happens based on how many feature points are detected and used in the matching process in SfM. \fig{fig:feature_point_increasing} shows the time history of the number of extracted feature points. As time passes, i.e., as the distance between the satellite and the camera gets closer and the proportion of the object in the image increases, more feature points are detected; therefore, the error of motion estimation is mitigated. Also, the periodic trend of the number of feature points is confirmed, which is reasoned by how many panels are visible in the image used. When the panel and the camera plane are parallel, which occurs periodically, the image includes only a single panel of the satellite, which decreases the features, resulting in less accuracy in the motion estimation.

\section{CONCLUSION}
\label{sec:conclusion}
In this paper, the novel algorithmic sequence for fully autonomous motion estimation and shape reconstruction of unknown shaped space debris was developed, combining the Structure from Motion (SfM) technique and several fundamental image processing libraries. This vision-based technique requires only a set of 2D images as input and, thus, does not depend on any pre-processes, additional hardware markers, and 3D data manipulation, which is practically preferable for the resource-constrained spacecraft's on-board computational capability. Such a contactless motion estimation is effective to observe the target motion towards the capture and additional services such as repairing, refueling, and deorbiting. The method has been successfully implemented and then applied to the single-axis rotation and the three-dimensional sinusoidal tumbling, which is typically observed in actual debris. The estimation result was compared with the ground truth, showcasing sufficient accuracy for 2D and 3D motions (relative error of the period $<$ 0.05\%).

In the future scope, the authors are willing to extend our approach to the mass centroid estimation based on the magnified analysis of the periodical rotation of the space debris, which is essential to apply for spacecraft where the geometrical and inertial centers do not match each other. Nonetheless, more advanced estimations of the inertial property, such as an inertia tensor, require contact-based information. For this, minimal force application and feedback analysis would be a solution.


\section*{Acknowledgment}
The authors would like to thank Andrew Price, Hiroki Okabayashi, Bas Bets, Rodrigo Silva, Yue Yu, and Swapnil Parekh for their help with the fabrication of the experimental apparatus and fruitful discussions.

\bibliography{./IEEEabrv,bibliography.bib}

\begin{thebibliography}{10}
\providecommand{\url}[1]{#1}
\csname url@rmstyle\endcsname
\providecommand{\newblock}{\relax}
\providecommand{\bibinfo}[2]{#2}
\providecommand\BIBentrySTDinterwordspacing{\spaceskip=0pt\relax}
\providecommand\BIBentryALTinterwordstretchfactor{4}
\providecommand\BIBentryALTinterwordspacing{\spaceskip=\fontdimen2\font plus
\BIBentryALTinterwordstretchfactor\fontdimen3\font minus \fontdimen4\font\relax}
\providecommand\BIBforeignlanguage[2]{{%
\expandafter\ifx\csname l@#1\endcsname\relax
\typeout{** WARNING: IEEEtran.bst: No hyphenation pattern has been}%
\typeout{** loaded for the language `#1'. Using the pattern for}%
\typeout{** the default language instead.}%
\else
\language=\csname l@#1\endcsname
\fi
#2}}

\bibitem{osoro2023sustainability}
O.~B. Osoro \emph{et~al.}, ``Sustainability assessment of low earth orbit (leo) satellite broadband mega-constellations,'' arXiv, 2023.

\bibitem{kessler1978collision}
D.~J. Kessler and B.~G. Cour-Palais, ``Collision frequency of artificial satellites: The creation of a debris belt,'' \emph{Journal of Geophysical Research: Space Physics}, vol.~83, no.~A6, pp. 2637--2646, 1978.

\bibitem{ets7}
N.~Inaba \emph{et~al.}, ``Autonomous satellite capture by a space robot: world first on-orbit experiment on a japanese robot satellite {ETS-VII},'' in \emph{Symp. Proc. IEEE Int. Conf. Robot. Automat.}, 2000, pp. 1169--1174.

\bibitem{darpaAstro}
A.~Ogilvie \emph{et~al.}, ``Autonomous satellite servicing using the orbital express demonstration manipulator system,'' in \emph{Proc. 9th Int. Symp. Artif. Intell., Robot. Automat. Spa. (i-SAIRAS)}, 2008, pp. 25--29.

\bibitem{uchida2024ISPARO}
A.~Uchida \emph{et~al.}, ``Space debris reliable capturing by a dual-arm orbital robot: Detumbling and caging,'' in \emph{Proc. the 2024 IEEE International Conference on Space Robotics (iSpaRo)}, 2024, pp. 194--201.

\bibitem{sensorFusion}
J.~Peng \emph{et~al.}, ``Pose measurement and motion estimation of space non-cooperative targets based on laser radar and stereo-vision fusion,'' \emph{IEEE Sensors Journal}, vol.~19, no.~8, pp. 3008--3019, 2019.

\bibitem{virtualStereoVision}
J.~Peng \emph{et~al.}, ``Virtual stereovision pose measurement of noncooperative space targets for a dual-arm space robot,'' \emph{IEEE Trans. Instrumentation and Measurement}, vol.~69, no.~1, pp. 76--88, 2020.

\bibitem{andrew}
A.~Price \emph{et~al.}, ``Render-to-real image dataset and cnn pose estimation for down-link restricted spacecraft missions,'' in \emph{Proc. 2023 IEEE Aerospace Conference}, 2023, pp. 1--11.

\bibitem{NNbasedPoseEst}
S.~Sharma \emph{et~al.}, ``Pose estimation for non-cooperative spacecraft rendezvous using convolutional neural networks,'' in \emph{Proc. IEEE Aerospace Conference}, 2018, pp. 1--12.

\bibitem{rembg}
\BIBentryALTinterwordspacing
D.~Gatis \emph{et~al.}, ``Rembg.'' [Online]. Available: \url{https://github.com/danielgatis/rembg}
\BIBentrySTDinterwordspacing

\bibitem{U2-Net}
X.~Qin \emph{et~al.}, ``{U2-Net}: Going deeper with nested u-structure for salient object detection,'' \emph{Pattern Recognit.}, vol. 106, no.~12, p. 107404, 2020.

\bibitem{pymatting}
T.~Germer \emph{et~al.}, ``{PyMatting}: A python library for alpha matting,'' \emph{Journal of Open Source Software}, vol.~~5, no. ~54, p. ~2481, 2020.

\bibitem{SfM}
M.~J. Westoby \emph{et~al.}, ``‘{S}tructure-from-{M}otion’ photogrammetry: A low-cost, effective tool for geoscience applications,'' \emph{Geomorphology}, vol. 179, pp. 300--314, 2012.

\bibitem{openmvg}
P.~Moulon \emph{et~al.}, ``Open{MVG}: Open multiple view geometry,'' in \emph{International Workshop on Reproducible Research in Pattern Recognition}, 2016, pp. 60--74.

\bibitem{opensfm}
\BIBentryALTinterwordspacing
``{OpenSfM}.'' [Online]. Available: \url{https://github.com/mapillary/OpenSfM}
\BIBentrySTDinterwordspacing

\bibitem{colmap}
J.~L. Sch\"{o}nberger \emph{et~al.}, ``Structure-from-motion revisited,'' in \emph{Proc. Conf. Computer Vision and Pattern Recognition (CVPR)}, 2016.

\bibitem{open3d}
Q.-Y. Zhou, J.~Park, and V.~Koltun, ``{Open3D}: {A} modern library for {3D} data processing,'' \emph{arXiv:1801.09847}, 2018.

\bibitem{Mariga_pyRANSAC-3D_2022}
\BIBentryALTinterwordspacing
L.~Mariga, ``{pyRANSAC-3D},'' 2022. [Online]. Available: \url{https://github.com/leomariga/pyRANSAC-3D}
\BIBentrySTDinterwordspacing

\bibitem{ransac}
A.~Martin \emph{et~al.}, ``Random sample consensus: a paradigm for model fitting with applications to image analysis and automated cartography,'' \emph{Commun. ACM}, vol.~24, no.~54, pp. 381--395, 1981.

\bibitem{uno2024lower}
K.~Uno \emph{et~al.}, ``Lower gravity demonstratable testbed for space robot experiments,'' in \emph{Proc. 2024 IEEE/SICE Int. Symp. Syst. Integrat. (SII)}, 2024, pp. 1183--1184.

\bibitem{nerf}
B.~Mildenhall \emph{et~al.}, ``{NeRF}: representing scenes as neural radiance fields for view synthesis,'' \emph{Communications of the ACM}, vol.~65, no.~1, pp. 99--106, 2021.

\end{thebibliography}

\end{document}